\documentclass[letterpaper, 10 pt, conference]{ieeeconf}  

\IEEEoverridecommandlockouts                              

\usepackage{listings}
\usepackage{minted}
\usepackage{amsmath}
\usepackage[capitalise]{cleveref}
\newcommand{\Llm}{\ensuremath{\mathcal{L}}}
\newcommand{\Tool}{\ensuremath{\Gamma}}
\newcommand{\Task}{\ensuremath{T}}
\newcommand{\Plan}{\ensuremath{\Pi_{\text{task}}}}
\usepackage{booktabs, makecell, subcaption}

\usepackage[most]{tcolorbox}
\usepackage{multirow}
\usepackage{makecell}
\usepackage{stfloats}

\usepackage{wrapfig}
\usepackage{algorithm}
\usepackage{algpseudocode}

\usepackage{makecell,multirow}  
\usepackage{caption}                     
\usepackage{adjustbox}                   
\usepackage{graphicx}                    

\usepackage{booktabs}
\usepackage[style=ieee, citestyle=numeric-comp, url=false, doi=false, natbib=true, mincitenames=1, maxcitenames=1]{biblatex}

\title{\LARGE \bf
General-Purpose Robotic Navigation via LVLM-Orchestrated Perception, Reasoning, and Acting
}

\author{
  Bernard Lange\textsuperscript{1,$\ast$}, Anil Yildiz\textsuperscript{1}, Mansur Arief\textsuperscript{1}, \\ Shehryar Khattak\textsuperscript{2}, Mykel Kochenderfer\textsuperscript{1}, Georgios Georgakis\textsuperscript{2}  \\ 
  \textsuperscript{1}Stanford University, \textsuperscript{2}Jet Propulsion Laboratory, California Institute of Technology
  \thanks{* Corresponding author.}
}

\begin{document}

\maketitle
\thispagestyle{empty}
\pagestyle{empty}

\begin{abstract}
Developing general-purpose navigation policies for unknown environments remains a core challenge in robotics. Most existing systems rely on task-specific neural networks and fixed information flows, limiting their generalizability. Large Vision–Language Models (LVLMs) offer a promising alternative by embedding human-like knowledge for reasoning and planning, but prior LVLM–robot integrations have largely depended on pre-mapped spaces, hard-coded representations, and rigid control logic.
We introduce the Agentic Robotic Navigation Architecture (ARNA), a general-purpose framework that equips an LVLM-based agent with a library of perception, reasoning, and navigation tools drawn from modern robotic stacks. At runtime, the agent autonomously defines and executes task-specific workflows that iteratively query modules, reason over multimodal inputs, and select navigation actions. This agentic formulation enables robust navigation and reasoning in previously unmapped environments, offering a new perspective on robotic stack design.
Evaluated in Habitat Lab on the HM-EQA benchmark, ARNA outperforms state-of-the-art EQA-specific approaches. Qualitative results on RxR and custom tasks further demonstrate its ability to generalize across a broad range of navigation challenges.
\end{abstract}
\section{Introduction}
\label{sec:intro}
Developing general-purpose navigation robots that can accomplish diverse tasks in unknown environments from natural language instructions remains a core challenge in robotics. This requires a robotic stack that integrates perception with flexible reasoning and decision-making, enabling robots to combine navigation, information retrieval, and logical inference as needed.

Traditional navigation systems rely on carefully engineered pipelines with tightly coupled perception, mapping, and planning modules. However, their task-specific policies and rigid control logic limit adaptability in unforeseen environments and tasks~\cite{wijayathunga2023challenges}. While end-to-end learning approaches have increased flexibility, they remain constrained by their training distribution and require supervision~\cite{levine2016end, shah2022gnm}. 

Recent advances in Large Vision-Language Models (LVLMs) offer a promising alternative. Pretrained on Internet-scale data, they embed rich semantic and commonsense knowledge and generalize impressively across domains from just a few examples~\cite{hurst2024gpt}. By enabling robots to reason over aggregated information and pursue abstract, language-defined goals, LVLMs provide an opportunity to fundamentally rethink the autonomy stack~\cite{rana2023sayplan, rajvanshi2024saynav, ahn2022can, brohan2023rt}.

\begin{figure}[t!]
    \centering
\includegraphics[width=0.95\columnwidth, trim={0 0 0 0}, clip]{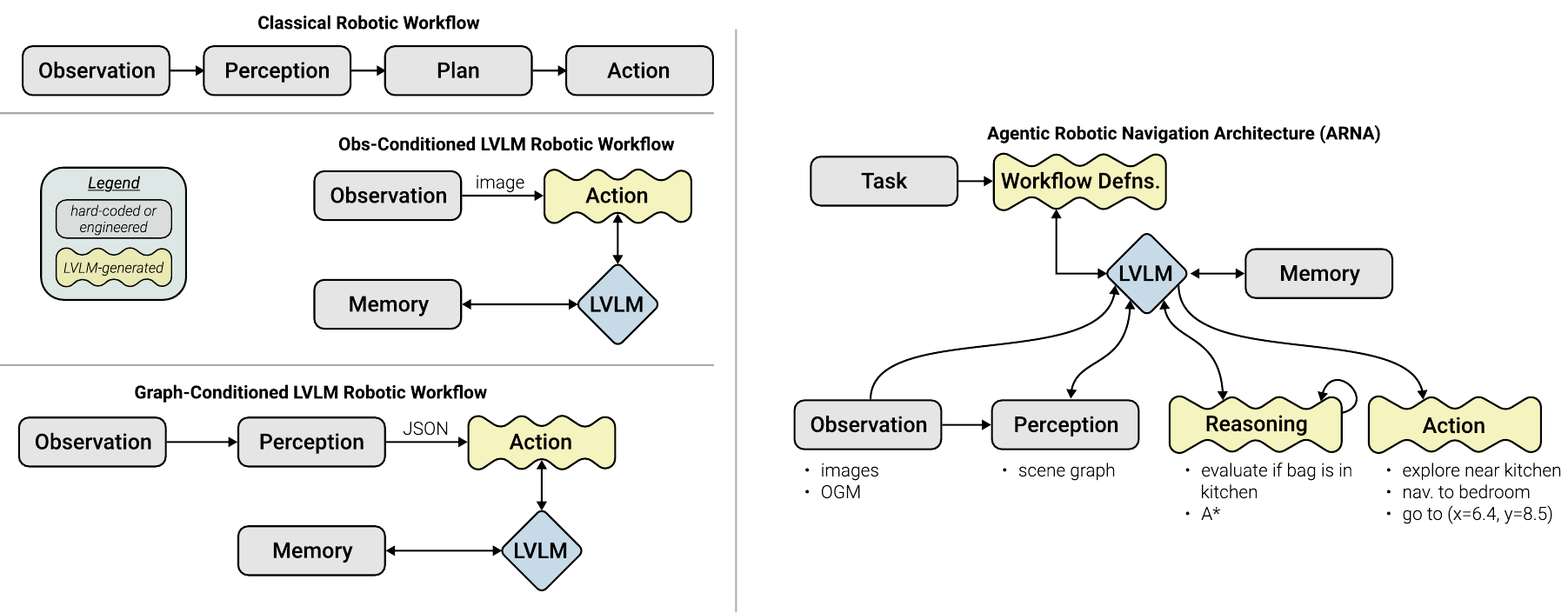}
    \caption{\small While classical robotic stacks (top left) and recent LVLM-based solutions (bottom left) follow fixed pipelines, ARNA (right) introduces a robotic agent that orchestrates perception, reasoning, and navigation tools to accomplish diverse, language-defined tasks.}
    \label{fig:overview_comparison}
    \vspace{-0.7cm}
\end{figure}

Yet, how to effectively integrate LVLMs into robotic stacks remains an open question. A common strategy uses them as expressive decision makers within predefined workflows, where the LVLM takes structured environment representations and outputs navigation actions. However, when prompted with raw observations to generate control inputs, LVLMs often exhibit short-sighted reasoning, hallucinations, or invalid action sequences due to limited spatial awareness~\cite{kamath2023s,chen2024spatialvlm,sun2024interactive,chen2025spatial} and insufficient physical understanding~\cite{ahn2022can,gao2024physically}. Consequently, prior work typically relies on simplified parameterizations—such as JSON-based scene graphs~\cite{rana2023sayplan,yin2024sg} or heuristically selected images~\cite{majumdar2023openeqa,exploreeqa2024}—and reduced action spaces in the form of sparse spatial nodes~\cite{majumdar2023openeqa,rana2023sayplan,yin2024sg,exploreeqa2024}.

Even if effective on simple tasks, such designs often yield myopic and brittle decisions, preventing generalization to novel or more complex tasks~\cite{majumdar2023openeqa}. They also fail to exploit the wealth of information and capabilities already available in modern robotic stacks, and overlook recent LVLM advances in agentic operation~\cite{yao2022react,schick2024toolformer,shinn2024reflexion}, where the model is cast as a controller in a perception–reasoning–action loop. In this agentic role, the LVLM can plan, invoke tools, and iteratively refine its reasoning, enabling pursuit of more complex tasks~\cite{mialon2023augmented,xi2023rise}. In this paper, we explore an agentic formulation of the robotic stack that addresses these limitations and augments existing systems with the reasoning and generalization capabilities of LVLMs.

We introduce the \textbf{A}gentic \textbf{R}obotic \textbf{N}avigation \textbf{A}rchitecture (\textbf{ARNA}), a framework that extends robotic stacks for task-agnostic navigation in unseen environments. At its core is an \textit{agentic robotic stack}, where an LVLM-based agent generates and executes task-specific workflows comprising information queries, reasoning functions, and navigation actions, as shown in \cref{fig:overview_comparison}. By accessing robotic capabilities through predefined tools, aggregating relevant information, and maintaining a multimodal internal memory, ARNA removes reliance on fixed state representations and rigid control logic. This design enables effective exploration and generalization to novel tasks, while leveraging the semantic, reasoning, and function-calling strengths of modern LVLMs. It further offers a blueprint for extending existing robotic stacks with LVLM-driven reasoning and generalization.
We benchmark ARNA in the Habitat Lab simulator~\cite{szot2021habitat} on HM-EQA~\cite{exploreeqa2024}, where it achieves state-of-the-art success rates and exploration efficiency. We further assess its generalization and limitations on a subset of RxR-CE~\cite{ku2020room} and on custom tasks, highlighting avenues for future improvement. Our approach lays the groundwork for generalizable navigation by extending robotic stacks with advanced agentic capabilities.
\section{Related Work}
Related work spans the integration of pre-trained foundational models into robotics and LVLM-based agents.

\textbf{Foundational Models.} Foundational models can be adapted to diverse tasks~\cite{bommasani2021opportunities,hurst2024gpt,radford2021learning,li2023blip}. Large Language Models (LLMs) capture human-like knowledge and reasoning abilities, while Large Vision–Language Models (LVLMs) extend this capability by grounding reasoning in visual input~\cite{hurst2024gpt}. Representation-learning-based vision–language models (VLMs) learn joint image–text embeddings~\cite{radford2021learning}.
They have been applied in robotics for navigation~\cite{rana2023sayplan,chen2024mapgpt,yin2024sg,yu2023l3mvn,yokoyama2024vlfm,shah2023lm}, manipulation~\cite{liang2023code,ahn2022can}, and embodied question answering~\cite{exploreeqa2024,majumdar2023openeqa}, even though they lack dedicated training for robotics.

\textbf{Zero-shot Integrations of Foundational Models.} Zero-shot applications of foundational models to navigation mainly explore I/O parameterizations for LVLMs~\cite{rana2023sayplan,yin2024sg,chen2024mapgpt,exploreeqa2024} or use VLM similarity scores for decision-making~\cite{shah2023lm,yokoyama2024vlfm}. LM-Nav~\cite{shah2023lm} combines language and vision models to follow textual instructions. SayPlan~\cite{rana2023sayplan} plans over a JSON-structured scene graph with semantic search and iterative replanning, while VLFM~\cite{yokoyama2024vlfm} generates language-grounded frontier maps to guide exploration. MapGPT~\cite{chen2024mapgpt} builds a topological map that is translated into prompts, and EuC~\cite{exploreeqa2024} directs exploration to high-value regions with calibrated confidence.
Beyond action generation, ConceptGraphs~\cite{gu2024conceptgraphs}, Embodied-RAG~\cite{xie2024embodied}, and 3D-Mem~\cite{yang20253d} propose customized memory and retrieval mechanisms for L(V)LM usage. 

\textbf{L(V)LM Agents.}
With their language interface, reasoning, and function-calling abilities, L(V)LMs have enabled agentic frameworks that plan, reason, and act over many steps to accomplish tasks~\cite{yao2022react,wu2023autogen,shinn2024reflexion,mialon2023augmented,xi2023rise,schick2024toolformer}. These frameworks demonstrate strong generalization across domains including web browsing~\cite{zheng2024gpt}, coding~\cite{zhang2024codeagent}, and gaming~\cite{wang2023voyager}.
In robotics, agentic approaches remain rare. To our knowledge, the only example is Code-as-Policies~\cite{liang2023code}, where an LLM generates code that accesses information and implements logic to solve manipulation tasks. Most navigation systems still rely on hand-crafted I/O representations within fixed control logic, limiting generalization. We propose a framework where the LVLM plans navigation and reasoning steps while invoking tools and information within the robotic stack.

\section{Preliminaries}
\label{sec:preliminaries}
We build on a traditional robotic stack that collects camera and depth measurements, stores semantic information in a scene graph, and navigates to any reachable observed location. Our framework augments this stack with an agentic layer. Below, we define the relevant terminology.

\subsection{Robotic Architecture}
\label{sec:robotic_architecture}
Traditional robotic systems follow the classic \emph{sense–plan–act} paradigm: sensor data are processed by a perception module, a planning layer selects the next action, and a control layer executes it. Interfaces, data flows, and task-specific reasoning logic are all handcrafted, forming what we refer to as the \emph{workflow}. Because both the workflow and module APIs are hardwired, these stacks struggle to adapt to new tasks or environments. End-to-end learned policies provide an alternative, but likewise generalize poorly to tasks or scenes not represented during training. LVLMs offer a promising direction, yet their integration into robotic systems remains an open question.

\subsection{Scene Graph}
\label{sec:scene_graph}
The scene graph provides a compact representation of information accumulated during robot exploration~\cite{hughes2024foundations}, combining perception with semantic reasoning. It is structured as a directed graph \(G = (V, E)\) with nodes \(V\) and edges \(E\). Nodes represent entities such as objects, traversable areas, or abstract spatial regions, together with their properties, while edges capture spatial or semantic relationships. The graph is organized into layers for objects, traversability, and region abstractions, and is augmented with a sensor layer that stores images, depth maps, and occupancy grids. Typically, the node attributes and connectivity rules are specified manually. Semantic classes may be assigned using closed-set perception methods or open-set methods without such constraints. We adopt the open-set formulation.

\subsection{LVLM Agent}
\label{sec:lvlm_agent} 
A Large Language Model (LLM) is a neural network trained on vast amounts of data sourced from the internet and other large-scale datasets~\cite{hurst2024gpt, hughes2024foundations}. 
Formally, an LLM processes a tokenized input sequence \( x = (x[1], \dots, x[n]) \) and models $p_\theta(x) = \prod_{i=1}^n p_\theta(x[i] \mid x[1], \dots, x[i-1])$.
A Large Vision Language Model (LVLM) extends this framework by conditioning text generation on image embeddings.
State-of-the-art LVLMs are fine-tuned to invoke functions from a provided toolset, enabling them to access external information through APIs. Their generalizability, remarkable reasoning and robust function-calling abilities have led to the development of L(V)LM-based agents, where the model acts as a controller for planning, memory, and acting~\cite{yao2022react,wu2023autogen,shinn2024reflexion,mialon2023augmented,xi2023rise,schick2024toolformer}.

\section{Agentic Robotic Navigation Architecture}
\label{sec:method}

\begin{figure*}[t!]
    \centering
\includegraphics[width=0.90\textwidth, trim={0 0 0 0}, clip]{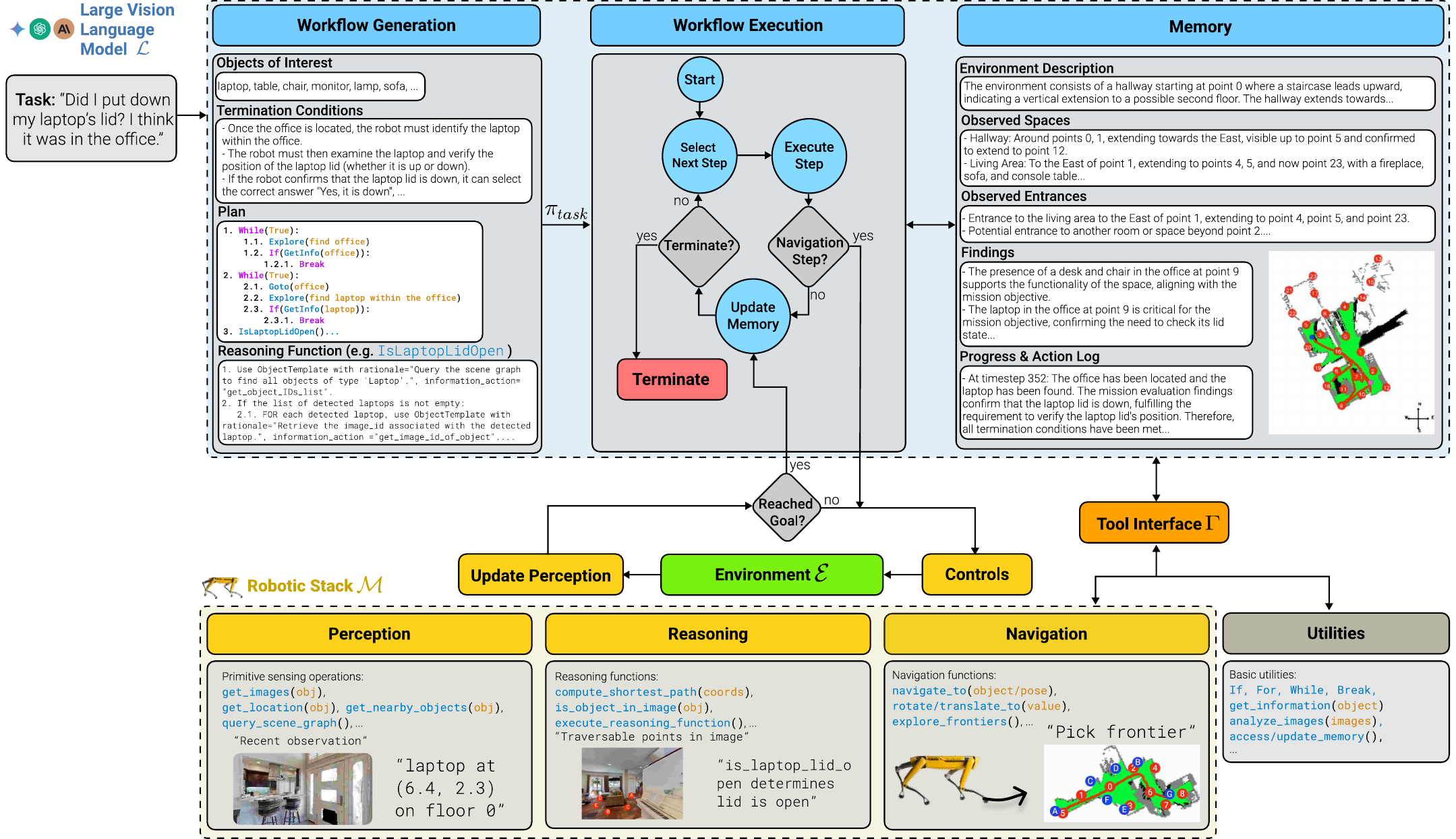} 
    \caption{\small
    ARNA is a navigation framework that equips an LVLM-based agent with a library of perception, reasoning, and navigation tools from a modern robotic stack. The agent autonomously defines and executes task-specific workflows that iteratively query modules, reason over multimodal inputs, choose navigation actions, and update its memory to fulfill any provided task.}
    \label{fig:detailed}
    \vspace{-0.5cm}
\end{figure*}

Traditional robotic architectures rely on fixed workflows and handcrafted representations, limiting generalization to new tasks. LVLMs offer rich semantic and reasoning capabilities, but existing work constrain them to simplified integrations. We propose the \textbf{A}gentic \textbf{R}obotic \textbf{N}avigation \textbf{A}rchitecture (\textbf{ARNA}), which embeds an LVLM agent into the robotic stack to enhance reasoning and enable generalization. ARNA generates and executes task-specific workflows—spanning exploration, information retrieval, reasoning, and navigation—by querying and invoking modules as needed (see \cref{fig:detailed}). This section presents the problem formulation along with the agent definition and its parts.

\subsection{Problem Formulation}
Given a task \(T\), the robot must achieve the objective through exploration, reasoning, and navigation in an unmapped environment \(\mathcal{E}\) that provides raw sensor observations \(o_t\) at each timestep \(t\). In a conventional robotic system \(\mathcal{S}\), this is accomplished through an expert-designed stack with a fixed workflow \(\pi_{\text{fixed}}\), i.e., a predefined flow of information, reasoning logic, and navigation steps. In ARNA, the challenge is to enable the agent to autonomously define and execute a task-dependent workflow \(\pi_{\text{task}}\) while leveraging the full capabilities of the robotic system.


\subsection{Agent Definition}
\label{sec:agent_definition} 
We model ARNA as the tuple \(\bigl(\mathcal{S},\,\Gamma,\,\mathcal{L},\,\Pi,\,\mathcal{M},\,T\bigr)\).
It augments the conventional stack \(\mathcal{S}\) by equipping an LVLM agent \(\mathcal{L}\) with read-and-execute access to all robotic modules through the tool interface \(\Gamma\).
Given a task \(T\), the agent generates and executes a task-specific workflow \(\pi_{\text{task}} \in \Pi\!\bigl(T,\,\Gamma(\mathcal{S}),\,\mathcal{L}\bigr)\), chaining tool calls and commands to gather information, reason, and issue navigation actions until the objective is achieved.
ARNA replaces the rigid pipeline \(\pi_{\text{fixed}}\) with a generated routine \(\pi_{\text{task}}\) that leverages existing robotic capabilities, removing the hard-coded workflows and representations typical of traditional stacks and prior LVLM integrations.
At every step of execution, ARNA maintains a memory \(\mathcal{M}\) that stores its findings, action history, and progress logs, ensuring continuity across steps. Every element of the ARNA tuple is managed by the LVLM using dedicated prompts that specify the expected behavior~\cite{hurst2024gpt}.

\subsection{Robotic Toolset}
ARNA accesses the robotic stack and collected information through a library of \emph{tools}, which the LVLM agent can invoke at runtime. These tools fall into three categories:

\textbf{Perception.} Tools that provide access to raw sensor measurements (e.g., \texttt{get\_images(timestep)}) and derived representations such as occupancy grids, or 3D scene-graph nodes (e.g., \texttt{query\_scene\_graph(label)}).

\textbf{Reasoning Utilities.} Tools that implement predefined or generated reasoning functions. Predefined utilities include shortest-path computation, visibility checks, reachability and proximity queries, and semantic labeling on images or grids. The agent can also generate custom functions (e.g., \texttt{is\_backpack\_near\_sofa()}) to capture repeated reasoning patterns or finer-grained logical subtasks. The generation process is described in the workflow section.

\textbf{Navigation \& Control.} Tools for movement and exploration, such as motion primitives and exploration routines. Examples include navigation commands (\texttt{goto(x,y,z)} with $A^*$, \texttt{rotate($\theta$)}), path following (\texttt{follow\_path(node)}), and LVLM-based frontier exploration (\texttt{explore\_frontiers()})~\cite{exploreeqa2024}.

These tools provide just-in-time access to information and core stack functionalities. With the exception of custom reasoning functions, all tools are manually defined and serve as the interface between the robotic stack and the LVLM agent. Arguments are generated by the LVLM based on the workflow and the environment state and progress encoded in memory. The API is designed to be extensible, allowing new algorithms to be integrated without modifying the agent definition. At inference time, the LVLM invokes tools through the interface $\Gamma$, supplying model-generated arguments (e.g., an open-set category for \texttt{query\_scene\_graph}) and issuing follow-up queries when additional context is required. Each tool includes comments describing its usage, which are provided to the LVLM as part of the prompt. For further details on tool usage, we refer to the OpenAI documentation\footnote{https://platform.openai.com/docs/guides/function-calling}.

\subsection{Memory Representation}
\label{subsec:memory}
\begin{figure*}[t!]
    \centering
\includegraphics[width=0.95\textwidth, trim={0 0 0 0}, clip]{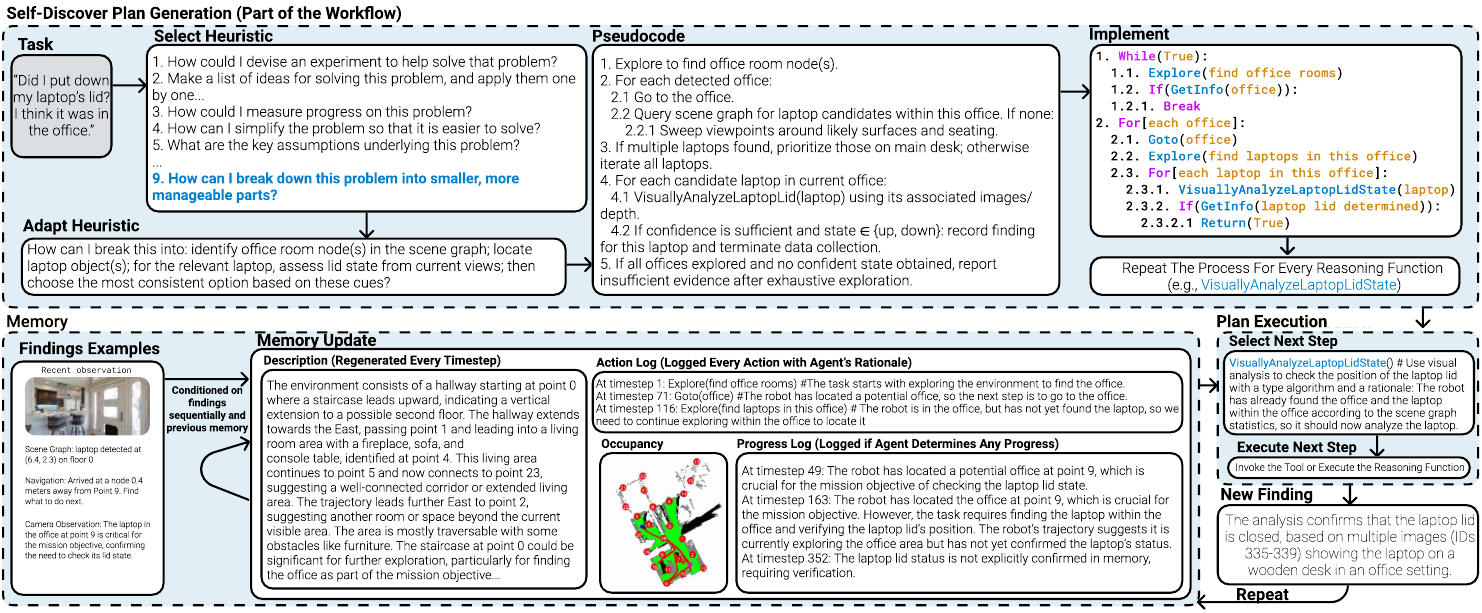}
    \caption{\small (Top) Visualization of plan generation with examples inspired by the self-discover approach~\cite{zhou2024self}. (Bottom) Visualization of the memory update process with examples. Both the plan and other workflow components, together with memory, are then used to guide plan execution, as illustrated. Each step is accompanied by dedicated prompts that describe the intended usage and examples for the LVLM.}
    \label{fig:workflow_memory}
    \vspace{-0.5cm}
\end{figure*}

A central component of the framework is its multimodal memory, which grounds the LVLM’s reasoning in an updatable representation, reducing redundant tool use and enabling effective progress tracking. It consists of: (1) a textual representation containing semantic and spatial descriptions, notable findings, and an automatically populated action log; and (2) a rasterized representation, a top-down occupancy grid annotated with landmarks corresponding to previously queried objects or sampled points within rooms or images, as illustrated in \cref{fig:detailed} and \cref{fig:workflow_memory}. Except for the action log, all elements of the memory remain under the LVLM’s direct control. In the textual representation, the agent is encouraged to reference the occupancy grid and its landmarks when producing spatial descriptions. The LVLM is conditioned on the memory at every step of workflow execution. After each step, the memory is incrementally updated by the LVLM with new findings and observations, as shown in \cref{fig:workflow_memory}.

\subsection{Workflow Definition, Generation, and Execution}
\label{subsec:wf_def}
Given a task, the available tools, and custom prompts, the agent generates a workflow \(\pi_{\text{task}}\) tailored to achieving the objective. This workflow specifies target objects for open-set perception, establishes multi-step plans, defines custom reasoning functions when needed, and sets termination conditions, as shown in \cref{fig:detailed} and \cref{fig:workflow_memory}. It consists of:

\begin{itemize}
    \item \textbf{Open-set Perception Initialization:} Identifies task-relevant objects to perceive based on the description.
    \item \textbf{Termination Conditions:} Defines the criteria for successful task completion and workflow termination.
    \item \textbf{Plan:} Outlines a high-level sequence of navigation, information-gathering, and reasoning steps. Reasoning may invoke predefined or dynamically generated functions for repeated or complex decisions, such as evaluating spatial relationships or interpreting scene layouts.
\end{itemize}

The workflow generation is compatible with various prompting techniques. We adopt a modified \textit{self-discover} approach~\cite{zhou2024self} for plan generation, which introduces structured intermediate steps to simplify the creation of complex plans that would otherwise require specialized prompting. The process consists of four stages: (1) selecting a heuristic from a predefined set of problem-solving strategies, (2) adapting this heuristic to the specific task, (3) generating pseudocode, and (4) implementing the plan using predefined tools (e.g., \texttt{get\_images(object)}, \texttt{goto(object)}), basic control-flow constructs (e.g., \texttt{if}, \texttt{while}), and dynamically generated reasoning functions (e.g., \texttt{is\_backpack\_near\_sofa()}). This procedure is also applied recursively when generating custom reasoning functions. See \cref{fig:workflow_memory} for an example. Open-set perception initialization and termination conditions are produced in a zero-shot manner. 

The agent then transitions to the \emph{execution} phase. At each iteration, the LVLM selects the next workflow step based on its progress as represented in memory. ARNA imposes no fixed ordering; order depends solely on the model’s current findings and reasoning, relying on the expressiveness of the underlying LVLM. Steps may be repeated or executed out of order if deemed necessary by the model. If a selected step specifies a navigation goal, control is passed to the motion stack, prompting the robot to move accordingly. Otherwise, the relevant tool is invoked, results are stored in memory, and the process continues. Generated reasoning functions are handled in the same manner. The workflow can be revised online as the agent explores the scene, but this yielded no tangible gains and only increased inference time, so we adopt it as a trade-off favoring efficiency.
The process is summarized in Algorithm~\ref{alg:arna-exec}, and shown in \cref{fig:detailed} and \cref{fig:workflow_memory}.

\begin{algorithm}[t]
\footnotesize
\caption{Agentic Robotic Navigation Architecture}
\label{alg:arna-exec}
\begin{algorithmic}[1]
\Require task description \Task, tool interface \Tool, LVLM agent \Llm

\State \Plan $\gets$ \Call{WorkflowGeneration}{\Llm, \Tool, \Task}             \Comment{generate workflow}
\State \textit{Mem} $\gets \emptyset$                                           \Comment{initialize memory}
\State \textit{ObsSeq} $\gets \emptyset$                                        \Comment{buffer for observations}

\While{\textbf{not} \Call{TERMINATE}{\Plan, \textit{Mem}}}
    \State \textit{step}   $\gets$ \Call{SelectStep}{\Llm, \Plan, \textit{Mem}}        \Comment{select next step from plan}
    \State \textit{result} $\gets$ \Call{ExecuteStep}{\Llm, \Tool, \textit{step}, \textit{Mem}} \Comment{execute the step}
    \State \Call{MemoryUpdate}{\Llm, \textit{Mem}, \textit{result}}                     \Comment{update memory}
    \If{\textit{result.type} = \textsc{NavGoal}}
        \State \textit{ObsSeq} $\gets \emptyset$                                        \Comment{reset observation buffer}
        \While{\textbf{not} \Call{Reached}{\textit{result.goal}}}
            \State \Call{NavigateStep}{\textit{result.goal}}                           \Comment{execute control actions}
            \State $o_t \gets$ \Call{GetObservation}{}                                  \Comment{sensor snapshot at time $t$}
            \State \Call{UpdatePerception}{$o_t$}                                            \Comment{update perception system}
            \State \textit{ObsSeq} $\gets \textit{ObsSeq} \cup \{o_t\}$                 \Comment{append to buffer}
        \EndWhile
        \State \Call{MemoryUpdate}{\Llm, \textit{Mem}, \textit{ObsSeq}}                 \Comment{update memory with observation sequence}
    \EndIf
\EndWhile
\end{algorithmic}
\end{algorithm}
\section{Experiment Setup}
All experiments are conducted in the Habitat Lab simulator~\cite{szot2021habitat}, a high-fidelity platform with photorealistic indoor environments from the HM3D~\cite{ramakrishnan2021hm3d} and MP3D~\cite{Matterport3D} datasets. We benchmark ARNA quantitatively against state-of-the-art methods on the HM-EQA benchmark~\cite{exploreeqa2024}. To further illustrate its generalizability, we qualitatively evaluate ARNA on selected scenes from RxR-CE~\cite{ku2020room}, as well as on custom tasks beyond standard benchmarks, presented in the video.

The scene graph is updated in real time using Grounding DINO~\cite{liu2023grounding}, while rooms are segmented with GPT-4o-mini~\cite{hurst2024gpt}.
For numerical evaluation, ARNA uses GPT-4o~\cite{hurst2024gpt} as the LVLM backbone. All experiments are conducted on an NVIDIA H100 GPU with a 64-core CPU. 

\subsection{Baselines}
We qualitatively evaluate ARNA on the HM-EQA benchmark~\cite{exploreeqa2024}, a multiple-choice question answering task in which a robot must explore an unseen environment to answer queries involving object identification, counting, spatial reasoning, and multi-goal navigation. We compare ARNA against state-of-the-art LVLM-based methods that differ in their conditioning and reasoning strategies: \emph{Explore-Until-Confident (EuC)}~\cite{exploreeqa2024}, \emph{OpenEQA}~\cite{majumdar2023openeqa}, and \emph{SayPlan}~\cite{rana2023sayplan}. While OpenEQA and SayPlan operate on pre-mapped environments, both EuC and ARNA actively explore the scene. For EuC, we evaluate two variants—one using GPT-4o~\cite{hurst2024gpt} and one using Prismatic~\cite{karamcheti2024prismatic}. Within the OpenEQA suite, we consider several configurations: \textsc{Blind}, \textsc{Scene-Graph}, \textsc{Scene-Graph w/ Captions}, \textsc{Frame Captions}, \textsc{Multi-Frame VLM (Random)}, and \textsc{Multi-Frame VLM (CLIP)}~\cite{radford2021learning}.

\begin{table}[t]
\vspace*{0pt}
\centering
\setlength{\tabcolsep}{1pt}
\footnotesize
\scalebox{0.9}{%
  \begin{tabular}{@{}lccr@{}}
    \toprule
    \textbf{Method} & \makecell{\textbf{Accuracy*}\\\textbf{(\%)}} 
                    & \makecell{\textbf{Mean Path}\\\textbf{Length (m, $\downarrow$)}}
                    & \makecell{\textbf{Mean Token}\\\textbf{Usage} ($\times10^3$, $\downarrow$)} \\
    \midrule
    \multicolumn{4}{l}{\textit{L(V)LM Conditioned with Premapped Information (GPT-4o)}} \\
    \midrule
    Blind                   & 40 / 60 & \multirow{7}{*}{N/A} & $0.16 \pm 0.01$ \\
    Scene-Graph             & 45 / 55 &                      & $0.18 \pm 0.01$ \\
    Scene-Graph w/ Captions & 46 / 54 &                      & $1.08 \pm 0.07$ \\
    Frame Captions          & 46 / 54 &                      & $2.83 \pm 0.02$ \\
    SayPlan w/ Vision       & 47 / 53 &                      & $13.8 \pm 1.12$ \\
    Multi-Frame VLM (Rand)  & 56 / 44 &                      & $4.05 \pm 0.01$ \\
    Multi-Frame VLM (Clip)  & 66 / 34 &                      & $4.04 \pm 0.01$ \\
    \midrule
    \multicolumn{4}{l}{\textit{L(V)LM Interacting in Environment}} \\
    \midrule
    EUC-Prismatic           & 40 / 60 & $313.07 \pm 64.08$   & N/A               \\
    EUC-GPT-4o-mini         & 38 / 62 & $70.89 \pm 3.75$     & $42.32 \pm 1.85$  \\
    EUC-GPT-4o              & 61 / 39 & $175.91 \pm 44.09$   & $42.72 \pm 9.13$  \\
    \midrule
    \multicolumn{4}{l}{\textit{L(V)LM Interacting in Environment (Ours)}} \\
    \midrule
    \textbf{ARNA (GPT-4o)}         & \textbf{77 / 14} 
                            & \textbf{16.55 $\pm$ 1.75} 
                            & $673.11 \pm 85.89$ \\
    \bottomrule
  \end{tabular}%
}
\caption{\small Results on the HM-EQA dataset. Accuracy is reported as Correct $(\uparrow)$ / Incorrect $(\downarrow)$. For ARNA results, values do not sum to 100, this is intentional.}
\label{tab:results}
\vspace{-0.7cm}
\end{table}%

\subsection{Ablation Study}
To assess the contributions of our navigation framework, we conduct ablations along two dimensions: (1) removing key representations such as the scene graph or occupancy grid, and (2) omitting all rasterization-based representations. In addition, we compare our self-discover-based workflow generation against a standard chain-of-thought approach.

\subsection{Evaluation Metrics}
We define the outcome of an experiment as \emph{success} if the agent answers correctly, \emph{failure} if the answer is incorrect, and \emph{inconclusive} if the agent does not complete the task within the allocated budget—either 500 reasoning steps in the workflow or a \$5 cost limit. 
Inconclusive runs are common in the agentic setting due to the dynamic nature of L(V)LM agents, which may continue running indefinitely. 

We benchmark using three metrics: (1) \textit{Accuracy}, the percentage of multiple-choice questions answered correctly; (2) \textit{Mean Path Length}, the average trajectory length (in meters), measuring exploration efficiency; and (3) \textit{Mean Token Usage}, the average number of tokens processed by the LVLM, which serves as a proxy for computational cost and runtime. Regarding the accuracy, we highlight that the success rate, failure rate, and inconclusive rate are reported separately and always sum to 1.

\section{Results and Analyses}
\label{sec:results}

\subsection{Baseline Comparison}
Our approach achieves substantial improvements on the HM-EQA benchmark over state-of-the-art task-specific models such as EuC, as well as SayPlan and general LVLM-based methods from the OpenEQA suite. Specifically, ARNA attains a correct/incorrect accuracy of 77\%/14\%, surpassing all baselines (see~\cref{tab:results}). The remaining 9\% corresponds to inconclusive trials, predominantly due to exceeding the allocated runtime or exploration budget, as detailed further in~\cref{sec:inconclusive}.

In terms of efficiency, ARNA achieves a mean path length of 16.55 meters, substantially shorter than all EuC-based policies (175.91, 70.89, and 313.07 meters for GPT-4o, GPT-4o-mini, and Prismatic, respectively). This notable reduction underscores ARNA's capability to explore effectively, minimize unnecessary navigation steps, and determine best termination timestep. All remaining baselines operate under the advantage of pre-mapped environments. However, the mean token usage for our model ($673.11 \times 10^3$) was substantially higher than all prior work, reflecting the increased computational demands of our reasoning agent. We highlight that the premapped baselines didn't need to spend any tokens to explore the environment.



\begin{figure*}[t]
    \centering
    \includegraphics[width=0.98\textwidth]{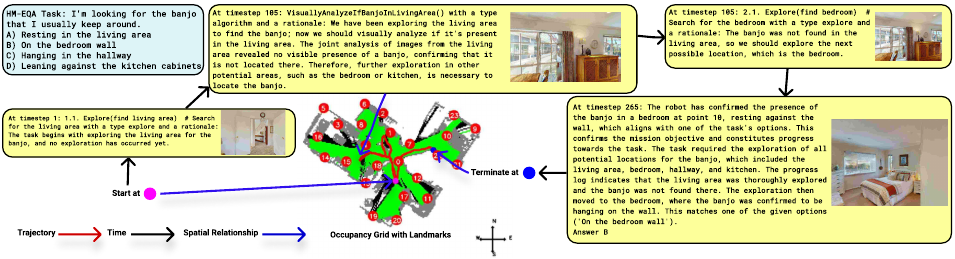}\\[1ex]
    \includegraphics[width=0.98\textwidth]{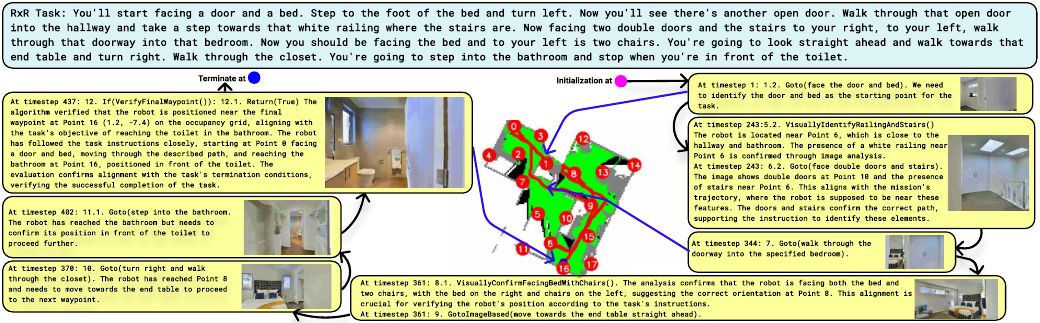}
    \caption{\small Examples on HM-EQA (top) and RxR (bottom) tasks, showing key decisions and findings with reference to the occupancy grids. The LVLM’s outputs provide insight into the agent’s reasoning behind these decisions. In the HM-EQA, the agent systematically explores the environment to visually confirm the presence of a banjo in a bedroom (top), while in RxR it navigates to locate a toilet (bottom).}
    \label{fig:example_result}
    \vspace{-0.5cm}
\end{figure*}

\subsection{Ablation Study}
The ablation study in \cref{tab:eqa_ablations} highlights the contributions of key components in our framework and the effectiveness of aggregating multiple input modalities. Removing the scene graph (NoSG) substantially reduces accuracy (41\% correct, 27\% incorrect) and increases the mean path length to 21.68 meters, underscoring the complementary role scene graphs play alongside image-based LVLM reasoning. 

Excluding the occupancy grid map (OGM) from memory (NoOGM) further decreases accuracy (33\% correct, 15\% incorrect), increases mean path length (24.3 meters), and raises the rate of inconclusive runs to 52\%. This demonstrates the critical role of the occupancy grid in enabling efficient exploration and spatial reasoning. Removing all rasterized representations (NoRP) results in a sharp drop in accuracy (10\% correct, 49\% incorrect), although the mean path length remains comparable to other ablations. While scene graphs and textual image descriptions, commonly used in prior work~\cite{rana2023sayplan, majumdar2023openeqa}, can guide navigation to specific locations, the absence of direct visual inputs severely limits the agent’s ability to answer questions accurately. These findings emphasize the importance of visual context: despite known limitations in spatial reasoning of LVLMs~\cite{chen2024spatialvlm, chen2025spatial}, visual modalities remain essential for robust performance, even when interpretations are occasionally imperfect.


We also evaluate a simplified workflow generation strategy based solely on chain-of-thought (CoT) prompting. This approach yields the lowest accuracy (7\% correct, 2\% incorrect) and the highest mean token usage, underscoring the importance of structured workflow generation. The step-by-step process in our method is crucial for effective decision-making and for efficiently managing and reasoning over the large set of available tools and information sources.

\begin{table}[bh]
  \centering
  \setlength{\tabcolsep}{2pt}
  \small
  \scalebox{0.95}{
    \begin{tabular}{@{}lccr@{}}
      \toprule
      \textbf{Method} & \makecell{\textbf{Accuracy*}\\\textbf{(\%)}} 
                      & \makecell{\textbf{Mean Path}\\\textbf{Length (m, $\downarrow$)}}
                      & \makecell{\textbf{Mean Token}\\\textbf{Usage} ($\times10^3$, $\downarrow$)} \\
      \midrule
      NoSG  & 41 / 27 & 21.68 $\pm$ 2.55  & 687.66 $\pm$ 60.70 \\
      NoOGM & 33 / 15 & 24.30 $\pm$ 1.78  & 612.86 $\pm$ 29.87 \\
      NoRP  & 10 / 49 & 20.46 $\pm$ 2.91  & 523.57 $\pm$ 90.80 \\
      CoT   & 7 / 2   & 32.91 $\pm$ 2.86  & 938.50 $\pm$ 87.37 \\
      \bottomrule
    \end{tabular}
  }
  \caption{\small Representations Ablation Study.}
  \label{tab:eqa_ablations}
\end{table}

Overall, the results show that our approach successfully integrates and reasons over multiple input modalities and tools to support efficient exploration, navigation, and decision-making. However, this improved effectiveness comes at the cost of higher computation per run than the baselines. 

\subsection{Inconclusive Runs}
\label{sec:inconclusive}
While L(V)LM agents have demonstrated impressive capabilities and generalization across numerous domains~\cite{mialon2023augmented,xi2023rise}, a major drawback remains their computational time and cost, which we analyze in \cref{tab:eqa_inconclusive}. The results highlight several failure modes observed across the ablations, offering insight into the limitations of each configuration and of the agentic setting more broadly. Inconclusive runs typically arise from (1) reaching the maximum number of environment steps (500), (2) exceeding the cost budget (maximum \$5 per experiment), or (3) incorrect function calls that terminate the run. Time and cost violations are primarily caused by inefficient exploration and suboptimal decision-making, whereas function call errors stem from mistakes in LVLM function invocation. All ablation models exhibit a marked increase in time-limit violations. In some cases, budget-triggered terminations occurred even when the task appeared close to completion, suggesting that additional resources could have enabled success.

\begin{table}[h]
  \centering
  \setlength{\tabcolsep}{2pt}
  \small
  \scalebox{0.95}{
    \begin{tabular}{@{}lccc@{}}
      \toprule
      \textbf{Method} & \makecell{\textbf{Time Limit}\\\textbf{(\%, $\downarrow$)}}
                      & \makecell{\textbf{Cost Limit}\\\textbf{(\%, $\downarrow$)}}
                      & \makecell{\textbf{Function Call}\\\textbf{Error (\%, $\downarrow$)}} \\
      \midrule
      \textbf{ARNA}   & 0  & 3 & 6  \\
      NoSG              & 27 & 5 & 0  \\
      NoOGM             & 44 & 2 & 6  \\
      NoRP              & 27 & 6 & 8  \\
      CoT               & 69 & 0 & 21 \\
      \bottomrule
    \end{tabular}
  }
  \caption{\small Breakdown of inconclusive agent runs.}
  \label{tab:eqa_inconclusive}
  \vspace{-0.5cm}
\end{table}

\subsection{RxR and Custom Tasks Experiments}
\label{sec:rxr}
We have shown that ARNA outperforms state-of-the-art approaches on EQA tasks, the most widely used benchmark for robotic navigation in continuous environments. Unlike prior L(V)LM methods tied to specific tasks, ARNA exposes the robot’s interface through a set of tools, constraining it only by the expressivity of the LVLM and its memory representation. This enables direct application to a broad range of tasks, which to the best of our knowledge has not been demonstrated before. To test this capability, we evaluate ARNA on a subset of RxR-CE~\cite{ku2020room}, a dataset with long navigation instructions and complex language cues. This benchmark requires advanced spatiotemporal reasoning, often involves ambiguous instructions, and was designed for end-to-end learning systems. On 30 episodes, our model achieved a 12\% success rate despite not being designed for this task—an encouraging result given its difficulty. To further demonstrate generalization, we deploy ARNA on custom tasks in unmapped environments, such as \textit{describing the spatial layout of the environment} or \textit{finding a good place for a new TV}, with examples shown in the video.

\subsection{Qualitative Evaluation}
\label{sec:qualitative_eval}
An example deployment of ARNA on HM-EQA and RxR tasks is shown in \cref{fig:example_result}, highlighting its key decisions and findings. In HM-EQA, ARNA systematically explores candidate locations until it visually confirms the presence of the target object (a banjo in a bedroom), at which point execution terminates. In RxR, it correctly follows the described trajectory and identifies the goal location. These examples illustrate ARNA’s ability to mirror human reasoning, enabling efficient and intuitive exploration and problem-solving.

\section{Limitations and Future Work}
\label{sec:limitations}
ARNA opens a promising avenue toward general-purpose robots by leveraging the rapid progress in natural language reasoning. Below, we outlined its current limitations.

\textbf{Computation, memory, and cost requirements.}
Interacting with L(V)LM agents is resource-intensive. ARNA requires frequent sequential API calls at each step, leading to runtimes of 30–60 minutes and costs of \$1–3 per task.

\textbf{No LVLM fine-tuning.}  
Our approach requires no fine-tuning, enabling true zero-shot deployment. However, it relies heavily on carefully crafted prompts, which may vary across LVLM backbones. Fine-tuning could help reduce this dependence and better align models with task objectives, thereby improving robustness and overall performance.


\textbf{Lack of plan verification and re-planning.}
Once an action plan is generated, ARNA can reorder steps or repeat completed ones if progress stalls or obstacles arise, but it does not revise the workflow definition itself. While enabling such dynamic updates is straightforward, in our experiments they failed to improve HM-EQA performance and substantially increased inference time. Nevertheless, workflow revision could prove valuable for tasks like RxR-CE~\cite{ku2020room}, which require re-planning due to their spatiotemporal instructions.

\textbf{Memory representation and verification.}  
Another limitation lies in memory representation: inaccuracies or insufficient expressivity can lead to incorrect decisions. More principled methods for verifying and updating memory are an important direction for future research~\cite{xie2024embodied,yang20253d}.

\textbf{Integration with advanced modules.}  
This paper explores an agentic formulation of the robotic stack using simple module implementations and straightforward prompting. The framework could be strengthened by integrating more advanced components, including richer scene graph representations~\cite{gu2024conceptgraphs,yang20253d}, retrieval-augmented generation systems~\cite{xie2024embodied}, and calibrated confidence metrics for critical decisions~\cite{exploreeqa2024}. 

\textbf{Agent brittleness.}  
L(V)LM agents rely on sequences of decisions to succeed, which demands capable models. Even with these, careful prompt design is necessary to ensure the LVLM interprets the intent and the signature of each tool within the framework. At present, only the most expressive models can reliably support our framework; smaller backbones, such as GPT-4o-mini, proved unsuccessful. Other models, such as GPT-5, perform well but often require adapted prompting due to training-induced behavioral differences. Development of agentic robotic stack could become a benchmark for LVLM agents in embodied AI setting. 
\section{Conclusion}
We introduced the \textbf{A}gentic \textbf{R}obotic \textbf{N}avigation \textbf{A}rchitecture (\textbf{ARNA}), a general-purpose framework that integrates LVLM agents into the robotic stack to address the limited generalization of prior methods. ARNA generates and executes task-specific workflows that reason over multimodal inputs, issue navigation actions, and query information as needed. On HM-EQA, ARNA achieved state-of-the-art accuracy and efficiency, while qualitative results on RxR and custom tasks demonstrated generalization. These findings highlight the viability of ARNA’s agentic approach, paving the way for scalable, general-purpose robots.

\section*{ACKNOWLEDGMENT}

The research was carried out at the Jet Propulsion Laboratory, California Institute of Technology, under a contract with the National Aeronautics and Space Administration (80NM0018D0004).
\copyright 2025. All rights reserved.


{\small
\printbibliography
}

\end{document}